\documentclass[11pt]{article}
\usepackage[LTH,T1]{fontenc}
\usepackage[utf8]{inputenc}
\DeclareUnicodeCharacter{0E4E}{\textyamakkan}
\DeclareUnicodeCharacter{0E4F}{\textfongmun}
\DeclareUnicodeCharacter{0E5A}{\textangkhankhu}
\DeclareUnicodeCharacter{0E5B}{\textkhomut}
\DeclareUnicodeCharacter{0E01}{\thaiKoKai}
\DeclareUnicodeCharacter{0E02}{\thaiKhoKhai}
\DeclareUnicodeCharacter{0E03}{\thaiKhoKhuat}
\DeclareUnicodeCharacter{0E04}{\thaiKhoKhwai}
\DeclareUnicodeCharacter{0E05}{\thaiKhoKhon}
\DeclareUnicodeCharacter{0E06}{\thaiKhoRakhang}
\DeclareUnicodeCharacter{0E07}{\thaiNgoNgu}
\DeclareUnicodeCharacter{0E08}{\thaiChoChan}
\DeclareUnicodeCharacter{0E09}{\thaiChoChing}
\DeclareUnicodeCharacter{0E0A}{\thaiChoChang}
\DeclareUnicodeCharacter{0E0B}{\thaiSoSo}
\DeclareUnicodeCharacter{0E0C}{\thaiChoChoe}
\DeclareUnicodeCharacter{0E0D}{\thaiYoYing}
\DeclareUnicodeCharacter{0E0E}{\thaiDoChada}
\DeclareUnicodeCharacter{0E0F}{\thaiToPatak}
\DeclareUnicodeCharacter{0E10}{\thaiThoThan}
\DeclareUnicodeCharacter{0E11}{\thaiThoNangmontho}
\DeclareUnicodeCharacter{0E12}{\thaiThoPhuthao}
\DeclareUnicodeCharacter{0E13}{\thaiNoNen}
\DeclareUnicodeCharacter{0E14}{\thaiDoDek}
\DeclareUnicodeCharacter{0E15}{\thaiToTao}
\DeclareUnicodeCharacter{0E16}{\thaiThoThung}
\DeclareUnicodeCharacter{0E17}{\thaiThoThahan}
\DeclareUnicodeCharacter{0E18}{\thaiThoThong}
\DeclareUnicodeCharacter{0E19}{\thaiNoNu}
\DeclareUnicodeCharacter{0E1A}{\thaiBoBaimai}
\DeclareUnicodeCharacter{0E1B}{\thaiPoPla}
\DeclareUnicodeCharacter{0E1C}{\thaiPhoPhung}
\DeclareUnicodeCharacter{0E1D}{\thaiFoFa}
\DeclareUnicodeCharacter{0E1E}{\thaiPhoPhan}
\DeclareUnicodeCharacter{0E1F}{\thaiFoFan}
\DeclareUnicodeCharacter{0E20}{\thaiPhoSamphao}
\DeclareUnicodeCharacter{0E21}{\thaiMoMa}
\DeclareUnicodeCharacter{0E22}{\thaiYoYak}
\DeclareUnicodeCharacter{0E23}{\thaiRoRua}
\DeclareUnicodeCharacter{0E24}{\thaiRu}
\DeclareUnicodeCharacter{0E25}{\thaiLoLing}
\DeclareUnicodeCharacter{0E26}{\thaiLu}
\DeclareUnicodeCharacter{0E27}{\thaiWoWaen}
\DeclareUnicodeCharacter{0E28}{\thaiSoSala}
\DeclareUnicodeCharacter{0E29}{\thaiSoRusi}
\DeclareUnicodeCharacter{0E2A}{\thaiSoSua}
\DeclareUnicodeCharacter{0E2B}{\thaiHoHip}
\DeclareUnicodeCharacter{0E2C}{\thaiLoChula}
\DeclareUnicodeCharacter{0E2D}{\thaiOAng}
\DeclareUnicodeCharacter{0E2E}{\thaiHoNokhuk}
\DeclareUnicodeCharacter{0E2F}{\thaiPaiyannoi}
\DeclareUnicodeCharacter{0E30}{\thaiSaraA}
\DeclareUnicodeCharacter{0E31}{\thaiMaiHanakat}
\DeclareUnicodeCharacter{0E32}{\thaiSaraAa}
\DeclareUnicodeCharacter{0E33}{\thaiSaraAm}
\DeclareUnicodeCharacter{0E34}{\thaiSaraI}
\DeclareUnicodeCharacter{0E35}{\thaiSaraIi}
\DeclareUnicodeCharacter{0E36}{\thaiSaraUe}
\DeclareUnicodeCharacter{0E37}{\thaiSaraUee}
\DeclareUnicodeCharacter{0E38}{\thaiSaraU}
\DeclareUnicodeCharacter{0E39}{\thaiSaraUu}
\DeclareUnicodeCharacter{0E3A}{\thaiPhinthu}
\DeclareUnicodeCharacter{0E3F}{\textbaht}
\DeclareUnicodeCharacter{0E40}{\thaiSaraE}
\DeclareUnicodeCharacter{0E41}{\thaiSaraAe}
\DeclareUnicodeCharacter{0E42}{\thaiSaraO}
\DeclareUnicodeCharacter{0E43}{\thaiSaraAiMaimuan}
\DeclareUnicodeCharacter{0E44}{\thaiSaraAiMaimalai}
\DeclareUnicodeCharacter{0E45}{\thaiLakkhangyao}
\DeclareUnicodeCharacter{0E46}{\thaiMaiyamok}
\DeclareUnicodeCharacter{0E47}{\thaiMaitaikhu}
\DeclareUnicodeCharacter{0E48}{\thaiMaiEk}
\DeclareUnicodeCharacter{0E49}{\thaiMaiTho}
\DeclareUnicodeCharacter{0E4A}{\thaiMaiTri}
\DeclareUnicodeCharacter{0E4B}{\thaiMaiChattawa}
\DeclareUnicodeCharacter{0E4C}{\thaiThanthakhat}
\DeclareUnicodeCharacter{0E4D}{\thaiNikhahit}
\DeclareUnicodeCharacter{0E4E}{\thaiYamakkan}
\DeclareUnicodeCharacter{0E4F}{\thaiFongman}
\DeclareUnicodeCharacter{0E50}{\thaizero}
\DeclareUnicodeCharacter{0E51}{\thaione}
\DeclareUnicodeCharacter{0E52}{\thaitwo}
\DeclareUnicodeCharacter{0E53}{\thaithree}
\DeclareUnicodeCharacter{0E54}{\thaifour}
\DeclareUnicodeCharacter{0E55}{\thaifive}
\DeclareUnicodeCharacter{0E56}{\thaisix}
\DeclareUnicodeCharacter{0E57}{\thaiseven}
\DeclareUnicodeCharacter{0E58}{\thaieight}
\DeclareUnicodeCharacter{0E59}{\thainine}
\DeclareUnicodeCharacter{0E5A}{\thaiAngkhankhu}
\DeclareUnicodeCharacter{0E5B}{\thaiKhomut}

\usepackage{fonts-tlwg}
\DeclareRobustCommand{\thaifont}{\thaitext}

\usepackage[preprint]{acl}
\usepackage{tgtermes}
\usepackage{latexsym}
\usepackage{microtype}
\IfFileExists{inconsolata.sty}{\usepackage{inconsolata}}{}
\usepackage{graphicx}
\usepackage{placeins}
\usepackage{array}
\usepackage{booktabs}
\newcolumntype{L}[1]{>{\raggedright\arraybackslash}p{#1}}
\usepackage{multirow}
\usepackage{url}
\usepackage{tikz}
\usepackage{amsmath}
\usetikzlibrary{positioning, arrows.meta, shapes.geometric, fit, calc, backgrounds}
\usepackage{tikz-dependency}
\pgfsetlayers{background,depgroups,main}

\title{ThaiTrees: Thai Syntactic Dependency Trees Across Domains}

\author{Attapol T. Rutherford and Papatchol Thientong \\
  Department of Linguistics \\
  Chulalongkorn University}

\begin{document}
\maketitle

\begin{abstract}
Studying syntactic patterns in naturally occurring language requires a large parsed corpus, but manual annotation is costly and difficult to scale. Thai has a manually annotated dependency treebank for training and evaluating parsers, but lacks a large automatically parsed corpus for quantitative syntactic research. We present ThaiTrees, a 342M-token corpus drawn from news, Wikipedia, spoken transcripts, and social media. We develop a reproducible pipeline for cleaning, processing, and parsing Thai text under the Universal Dependencies framework. The resulting corpus makes grammatical relations searchable and supports the study of syntactic distributions. We release a frequency lexicon and CoNLL-U parses in machine-readable formats suitable for both AI-assisted and conventional programmatic analysis.
\end{abstract}

\section{Introduction}

Studying syntactic patterns in naturally occurring language requires a corpus
large enough to yield stable frequency estimates and support statistical
inference.  A treebank or other parsed corpus makes grammatical relations
queryable, enabling analyses of syntactic distributions, valency patterns, and alternations \citep{
lehmann-schneider-2013-bnc}.  However, manual syntactic annotation (or treebanking) requires
rare trained annotators and is costly to create and maintain at scale
\citep{marcus-etal-1993-penn}, so they are mainly used for training automatic parsers. The parser can be applied automatically to a much larger corpus.  Building
such automatically parsed corpora is consequently important for extending
quantitative syntactic research beyond the limited size of gold-standard
annotation.

Dependency grammar and Universal Dependency provide a convenient representation for computational
analysis because it encodes syntax directly as labeled head-dependent relations
between words \citep{de-marneffe-etal-2021-universal,nivre-etal-2017-universal}.  In a basic dependency tree, words are the nodes and
head--dependent relations are the arcs; unlike phrase-structure trees, no
intermediate constituent nodes need be introduced.  The representation remains
a rooted tree, but the fixed one-node-per-word structure means that parsing
connects existing lexical nodes directly 
\citep{nivre-2010-dependency}.  Automatically parsed dependency corpora can
therefore be searched for grammatical collocations and used to examine constructional alternations such as the
active--passive, verb--prepositional-phrase, and dative alternations
\citep{uhrig-etal-2018-collocation,lehmann-schneider-2013-bnc}.  At web
scale, such corpora have also supported syntax-based distributional models,
open information extraction, and question answering
\citep{panchenko-etal-2018-depcc}.

Thai already has a manually annotated dependency treebank: Thai-TUD contains over
3{,}600 sentences and provides a foundation for training and evaluating Thai
parsers \citep{sriwirote-etal-2024-tud}.  It does not, however, provide the
large automatically parsed corpus needed for corpus-scale syntactic analysis.
We therefore create ThaiTrees, a 342M-token corpus drawn from news,
Wikipedia, spoken transcripts, and social media.  Its sentences are parsed
automatically under the Universal Dependencies framework.  We develop a
reproducible pipeline for cleaning, processing, and parsing Thai text, and
release the frequency lexicon and CoNLL-U parses in machine-readable formats suitable both for AI-assisted and conventional programmatic analysis. \footnote{The corpus is available at \url{https://github.com/nlp-chula/thaitrees}}

\section{Related Work}

Thai resources provide gold-standard annotation at a scale suited to model
development.  UD Thai-TUD contains 3{,}627 manually annotated dependency
trees (77{,}215 tokens) drawn from the Thai National Corpus and Thai
Wikipedia \citep{sriwirote-etal-2024-tud}; we use a dependency parser trained
on it.  ORCHID provides manually checked sentence boundaries, word
segmentation, and POS tags for technical prose
\citep{charoenporn-etal-1997-orchid}; LST20 provides segmentation, POS,
named entities, and clause and sentence boundaries for 3.2\,M words of news
\citep{boonkwan-etal-2020-lst20}.  The Thai National Corpus supplies a
general written reference corpus and frequency interface; the 2009 progress
report documented 14\,M processed words and described collection constraints
from copyright clearance \citep{aroonmanakun-2007-creating,
aroonmanakun-etal-2009-progress}.  These resources are valuable for
training and evaluating models, but their scale, single-register coverage,
or lack of dependency annotation limits the corpus-linguistic claims they
can support.  ThaiTrees therefore adds a much larger, four-domain corpus for
distributional analysis, while relying on Thai-TUD as the gold-standard
source for parser training.

Large corpora with automatic dependency parses already support
corpus-linguistic research in English and other languages.  Sketch Engine,
for example, is designed to query dependency-parsed corpora and derive
grammatical-relation summaries \citep{kilgarriff-etal-2014-sketch};
its preloaded collections include English pukWaC, parsed with MaltParser
\citep{sketch-engine-corpus-list}.  The earlier WaCky web corpora provided
large, automatically linguistically processed resources for English, German,
and Italian \citep{baroni-etal-2009-wacky}, and the related TenTen family
extends this web-corpus approach across languages \citep{jakubicek-etal-2013-tenten}.
Thai currently lacks a comparably broad corpus with automatically produced,
queryable dependency analyses.  

Because Thai lacks explicit word and sentence boundaries, most processing
stages rely on machine-learning models tuned to Thai data.  AttaCut
\citep{chormai-etal-2020-syllable} performs word segmentation with a
convolutional neural network trained on annotated Thai syllable/word boundaries.  PyThaiNLP
and its CRFcut sentence segmenter \citep{phatthiyaphaibun-etal-2023-pythainlp,
chumpolsathien-2020-crfcut} use CRF-backed sentence-boundary models (with
internal newmm features) benchmarked on corpora such as ORCHID and TED
transcripts.  AttaParse 1.0 \citep{sriwirote-etal-2024-tud} wraps Stanza's
graph-based UD parser \citep{qi-etal-2020-stanza} and parses pre-tokenized
input using models trained on UD Thai-TUD.  A PhayaThaiBERT transformer,
fine-tuned on UD Thai-TUD, supplies UPOS tags that overwrite the parser's
column \citep{sriwirote-etal-2024-phayathaibert}.  These ML-based tools
perform decently well on in-domain material, and together show that Thai
has a wide array of automatic linguistic annotation tools available for use.

\section{ThaiTrees: Corpus Construction}

\subsection{Data Sources}

\begin{table*}[t]
\centering
\small
\begin{tabular}{lrrrl}
\toprule
\textbf{Domain} & \textbf{Documents} & \textbf{Tokens} & \textbf{Sentences} & \textbf{Source} \\
\midrule
News         & 100{,}846 &  52{,}641{,}127 &    897{,}348 & ThaiPBS website \\
Wikipedia    & 175{,}069 & 101{,}378{,}583 &  1{,}303{,}443 & Thai Wikipedia (MediaWiki random sampling) \\
Spoken       &   2{,}505 &  30{,}238{,}809 &    266{,}944 & YouTube transcripts (7+ channels) \\
Social Media &  87{,}700 & 157{,}708{,}614 &  2{,}052{,}040 & Wisesight + Pantip datasets \\
\midrule
\textbf{Total} & \textbf{366{,}120} & \textbf{341{,}967{,}133} & \textbf{4{,}519{,}775} & \\
\bottomrule
\end{tabular}
\caption{The four sub-corpora that make up ThaiTrees.}
\label{tab:corpus-overview}
\end{table*}


The four domains are journalistic prose (news), encyclopedic prose (Wikipedia),
conversational spoken language (YouTube podcasts, livestreams, and film
transcripts), and informal written conversation (online forum and
sentiment-dataset posts).  The Spoken sub-corpus has fewer documents but
each is roughly an order of magnitude longer, because each document is the
transcript of one long recording (Table~\ref{tab:corpus-overview}).

The news sub-corpus consists of over 100,000 articles from the ThaiPBS
public-broadcaster website. We only include articles that are made available publicly on the website in December 2025. 

Thai Wikipedia articles were sampled via MediaWiki's random-article
endpoint in batches of 500, namespace 0, with disambiguation pages excluded
and reference sections truncated. Sampling was done in December 2025.

Spoken-language transcripts were retrieved from YouTube through the
youtube-transcript.io API.  We manually select the channels that include manual transcription so that we get the highest quality transcription without using automatic speech recognition. The sources are the Thai PBS Podcast network, independent podcasts (bigboung, BeSider), film subtitles, Prachatai political-commentary livestreams, and SaltymanTH livestreams.

The social-media sub-corpus combines the Wisesight sentiment dataset
and Pantip forum used in training large language models such as WangchanBERTa
\citep{lowphansirikul-etal-2021-wangchanberta} and PhayaThaiBERT
\citep{sriwirote-etal-2024-phayathaibert}.

\subsection{Processing Pipeline}
\label{sec:pipeline}


\begin{figure*}[t]
\centering
\begin{tikzpicture}[
  node distance=4mm,
  stage/.style={
    rectangle, rounded corners=2pt, draw, thick,
    minimum height=14mm, minimum width=24mm,
    align=center, inner sep=1.2mm, fill=black!4,
    font=\scriptsize
  },
  io/.style={
    rectangle, dashed, draw, thick,
    minimum height=14mm, minimum width=18mm,
    align=center, inner sep=1.2mm,
    font=\scriptsize\itshape
  },
  branch label/.style={
    font=\tiny\itshape\sffamily, gray
  },
  flow/.style={-{Stealth[length=2.5mm]}, thick}
]

\node[io] (input) {raw Thai\\ text};

\node[stage, right=10mm of input, yshift=-9mm] (crfcut)
   {\textbf{CRFcut}\\[0.3ex] sentence\\ segmentation};
\node[stage, right=of crfcut] (attacut2)
   {\textbf{AttaCut}\\[0.3ex] tokenization\\ (per sentence)};
\node[stage, right=of attacut2] (attaparse)
   {\textbf{AttaParse 1.0}\\[0.3ex] dependency\\ parsing};
\node[stage, right=of attaparse] (pos)
   {\textbf{PhayaThaiBERT}\\[0.3ex] POS\\ overwrite};
\node[io, right=of pos] (conllu) {CoNLL-U};

\node[stage, right=10mm of input, yshift=9mm] (attacut1)
   {\textbf{AttaCut}\\[0.3ex] tokenization\\ (whole document)};
\node[io, right=of attacut1] (lexicon)
   {frequency\\ lexicon};

\node[branch label, above=1mm of attacut1.north west, anchor=south west]
   {lexicon branch};
\node[branch label, below=1mm of crfcut.south west, anchor=north west]
   {parsing branch};

\draw[flow] (input.east) -- (attacut1.west);
\draw[flow] (input.east) -- (crfcut.west);

\draw[flow] (attacut1.east) -- (lexicon.west);

\draw[flow] (crfcut.east) -- (attacut2.west);
\draw[flow] (attacut2.east) -- (attaparse.west);
\draw[flow] (attaparse.east) -- (pos.west);
\draw[flow] (pos.east) -- (conllu.west);

\end{tikzpicture}
\caption{The processing pipeline.  The raw text runs through two
branches: the lexicon branch (top) runs AttaCut over each whole
document, producing the frequency lexicon; the parsing branch
(bottom) runs CRFcut for sentence segmentation, re-applies AttaCut to
each sentence, parses with AttaParse 1.0
(\texttt{tokenize\_pretokenized=True}), and overwrites the POS column
with PhayaThaiBERT.}
\label{fig:pipeline}
\end{figure*}
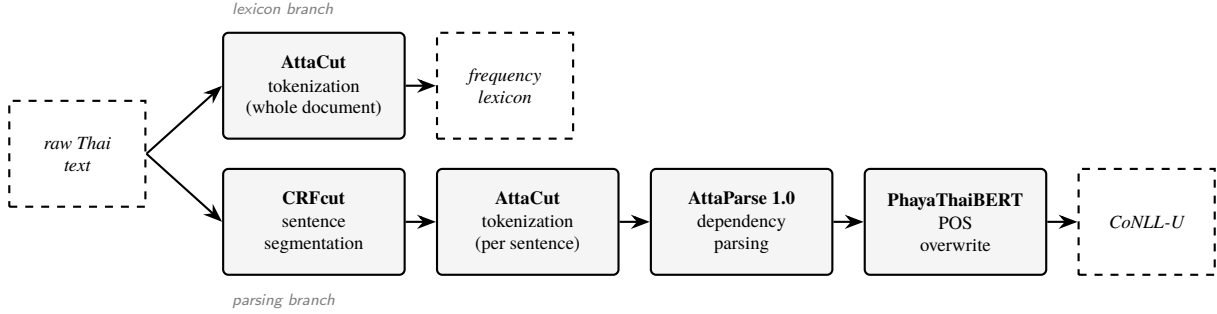

The lexicon and parsed corpus are produced in separate branches that
apply word segmentation to different units of text: whole documents for the lexicon
and individual sentences for parsing (Figure~\ref{fig:pipeline}).  The resulting token boundaries can differ,
so the two releases share document identifiers but not token identifiers. For each processing step that requires an machine-learning-based tool, we select the tools that achieve state-of-the-art results on some Thai benchmark data except for Thai POS tagger (Table~\ref{tab:tool-bench}).

\subsubsection{Word segmentation}
We use a CNN-based state-of-the-art Thai word segmenter, AttaCut \citep{chormai-etal-2020-syllable}, to segment each whole document
for the lexicon branch. We store its output in Parquet, replacing the
\texttt{text} column with pipe-delimited tokens and adding a
\texttt{token\_count} column. Lexicon construction then
removes punctuation, pure-numeric tokens, emoji, whitespace-only tokens, and
forms whose total corpus frequency is at most five.  This filtering is applied
to the document-level AttaCut output, independently of the parsing branch.

\subsubsection{Sentence segmentation and tokenization}
\label{sec:two-tokenizations}
The parsing branch reads the raw text directly.  It collapses newlines
and runs of spaces to a single space and caps social-media documents at
500{,}000 characters before sentence segmentation; this truncates 149 long,
concatenated documents.  PyThaiNLP's CRFcut
\citep{phatthiyaphaibun-etal-2023-pythainlp,chumpolsathien-2020-crfcut}
then returns sentence strings.  CRFcut uses dictionary-based word segmentation internally
to extract CRF features but discards those token boundaries when returning
the strings, but sentence segmentation itself does not remove text from its
normalized input.  This separate branch avoids carrying CRFcut's internal
token boundaries into the downstream word-segmentation step.
 
\subsubsection{Dependency parsing}
We parse with a graph-based neural dependency parser: AttaParse's
Thai-specific Stanza model uses PhayaThaiBERT contextual representations and
the no-POS configuration identified in Thai-TUD \citep{qi-etal-2020-stanza,sriwirote-etal-2024-tud}.  For each CRFcut-delimited
sentence, AttaParse receives the AttaCut tokens as pretokenized input and
predicts labeled head--dependent arcs while preserving those token boundaries.
Crucially, it requires neither POS tags nor lemmas: the POS and lemma
processors are disabled, \texttt{UPOS} and \texttt{XPOS} are set to
\texttt{.}, and \texttt{FEATS} to \texttt{\_} before parsing.  The released
\texttt{XPOS} column consequently remains \texttt{.}.

Whitespace-only tokens are discarded at the parser's pretokenized entry point.
Parsing output is written as 5\,M-token CoNLL-U chunks on document boundaries;
completed chunks are skipped and writes are atomic to allow resumption.  One
deterministic parser failure on Wikipedia chunk~16 is excluded from the final
release.  

\subsubsection{POS tagging and benchmark performance}

To our knowledge, no off-the-shelf Thai Universal POS (UPOS) tagger has been
benchmarked on Thai-TUD with a directly comparable held-out evaluation.  We
therefore train our own tagger on Thai-TUD \citep{sriwirote-etal-2024-tud},
fine-tuning a state-of-the-art encoder-only model for Thai, called PhayaThaiBERT \citep{sriwirote-etal-2024-phayathaibert}, for four
epochs with AdamW and a learning rate of $5\times10^{-5}$.  The tagger
achieves 90.64\% held-out accuracy and 81.34\% macro F1 across 15 UPOS
classes (Table~\ref{tab:tagger-config}).  Only the first sub-word prediction
for each word is retained for the prediction head. 

\begin{table}[t]
\centering
\footnotesize
\setlength{\tabcolsep}{5pt}
\begin{tabular}{@{}ll@{}}
\toprule
\textbf{Setting} & \textbf{Value} \\
\midrule
Base model & PhayaThaiBERT \\
Training data & UD Thai-TUD \\
Split (sentences) & 2{,}902 / 362 / 363 \\
Split (word tokens) & 62{,}011 / 7{,}521 / 7{,}683 \\
Optimizer & AdamW \\
$\beta_1,\ \beta_2,\ \epsilon$ & $0.9$,\ $0.999$,\ $10^{-8}$ \\
Learning rate & $5\times10^{-5}$ \\
Batch size & 16 \\
Epochs & 4 \\
Warmup ratio & 0.1 \\
Weight decay & 0.01 \\
Precision & fp16 mixed \\
Max sequence & 512 sub-words \\
\midrule
Test accuracy & 90.64\% \\
Test macro F1 & 81.34\% \\
\bottomrule
\end{tabular}
\caption{Fine-tuning configuration for the PhayaThaiBERT POS tagger
(\texttt{phayathaibert-thai-pos-tagger}).  The
best checkpoint was selected by held-out accuracy; macro F1 is over the 15
UPOS classes present in the treebank.}
\label{tab:tagger-config}
\end{table}

The predictions overwrite column 4 (UPOS) of the AttaParse CoNLL-U
output.  Since the dependency parser does not use POS features, this
changes no edges: tags and edges remain independent annotation layers
over the same tokens.

\begin{table*}[t]
\centering
\footnotesize
\setlength{\tabcolsep}{6pt}
\begin{tabular}{llll}
\toprule
\textbf{Stage} & \textbf{Tool} & \textbf{Benchmark} & \textbf{Source} \\
\midrule
Word seg.    & AttaCut       & 91\,\% WL-F1 (BEST)                                                & \citealp{chormai-etal-2020-syllable} \\
Sent. seg.   & CRFcut        & 87\,\% / 82\,\% acc. (ORCHID / TED)                                & \citealp{chumpolsathien-2020-crfcut} \\
Dep. parsing & AttaParse 1.0 & 86.4\,\% UAS / 76.6\,\% LAS (TUD test)\textsuperscript{$\dagger$}   & \citealp{sriwirote-etal-2024-tud} \\
POS tagging  & PhayaThaiBERT (ours) & 90.64\,\% / 81.34\,\% macro F1 (UD Thai-TUD)                & this work \\
\bottomrule
\end{tabular}
\caption{In-domain benchmarks for the four pipeline stages.
\textsuperscript{$\dagger$}AttaParse~1.0 = the no-POS graph-based
PhayaThaiBERT configuration (row GP) of
\citet{sriwirote-etal-2024-tud}, Table~3.}
\label{tab:tool-bench}
\end{table*}

\subsection{Released Artifacts}
\label{sec:layers}

Computing syntactic frequency at this scale produced three artifacts, which
we release together so the analyses in \S\ref{sec:wordfreq} and
\S\ref{sec:descriptive} can be reproduced.  Document identifiers are shared
across all three, and sentence identifiers across the parsed corpus, so results can be joined at the document level.  Token
identifiers do not align across the lexicon and the parsed corpus,
which index different tokenizations of the same raw text
(\S\ref{sec:pipeline}).

The raw text corpus contains 366{,}120 documents and
341{,}967{,}133 AttaCut-segmented tokens in Parquet format.  Its schema is \texttt{\{doc\_id, domain, text, token\_count\}}, where
\texttt{text} uses pipe delimiters to mark token boundaries.  Per-domain
files and a sentence-segmented variant (4{,}519{,}775 sentences) are
included.

The frequency lexicon contains 451{,}252 unique word forms and
255{,}434{,}917 filtered tokens, released as Parquet and SQLite.
Each row carries 34 columns.  A five-column core holds the form, its total
count, total rank, frequency per million, and cross-corpus document count.  Six
columns per domain give the count, rank, freq/M, document count, document
frequency, and IDF.  Five further columns record how many domains the form
appears in, a cross-domain flag, character length, IDF, and the dominant domain.
This artifact covers the full raw corpus through the document-level lexicon
branch.  39{,}465 word forms (8.7\%) appear in all four domains.

The dependency-parsed corpus contains 199{,}836{,}464 dependency edges
in 10-column CoNLL-U, chunked into 5\,M-token archives.  A per-pattern edge frequency table is released alongside; it lists
the 5{,}944 distinct patterns in the corpus (head POS, relation,
dependent POS) with per-domain counts.
The parsed corpus contains 203{,}892{,}200 tokens.

\section{Word Frequency}
\input{figures/fig_bump_top20}

\begin{table}[!t]
\centering
\footnotesize
\setlength{\tabcolsep}{4pt}
\begin{tabular}{rlrr}
\toprule
\textbf{Rank} & \textbf{Word} & \textbf{Count} & \textbf{Freq/M} \\
\midrule
 1 & {\thaifont ที่}   & 6{,}300{,}593 & 18{,}424.6 \\
 2 & {\thaifont ใน}   & 3{,}972{,}907 & 11{,}617.8 \\
 3 & {\thaifont การ}  & 3{,}928{,}739 & 11{,}488.6 \\
 4 & {\thaifont ไม่}   & 3{,}729{,}958 & 10{,}907.4 \\
 5 & {\thaifont ได้}   & 3{,}720{,}946 & 10{,}881.0 \\
 6 & {\thaifont มี}    & 3{,}655{,}798 & 10{,}690.5 \\
 7 & {\thaifont เป็น}  & 3{,}612{,}911 & 10{,}565.1 \\
 8 & {\thaifont จะ}   & 3{,}134{,}256 &  9{,}165.4 \\
 9 & {\thaifont และ}  & 3{,}015{,}246 &  8{,}817.4 \\
10 & {\thaifont ว่า}   & 2{,}869{,}729 &  8{,}391.8 \\
11 & {\thaifont ของ}  & 2{,}727{,}411 &  7{,}975.7 \\
12 & {\thaifont ก็}    & 2{,}625{,}733 &  7{,}678.3 \\
13 & {\thaifont ไป}   & 2{,}492{,}444 &  7{,}288.5 \\
14 & {\thaifont ให้}   & 2{,}458{,}395 &  7{,}189.0 \\
15 & {\thaifont มา}   & 2{,}368{,}727 &  6{,}926.8 \\
16 & {\thaifont นี้}    & 2{,}255{,}547 &  6{,}595.8 \\
17 & {\thaifont ๆ}    & 2{,}027{,}557 &  5{,}929.1 \\
18 & {\thaifont จาก}  & 1{,}791{,}281 &  5{,}238.2 \\
19 & {\thaifont คน}   & 1{,}714{,}923 &  5{,}014.9 \\
20 & {\thaifont แล้ว}  & 1{,}639{,}241 &  4{,}793.6 \\
\bottomrule
\end{tabular}
\caption{Top 20 words across the corpus by overall frequency per million.}
\label{tab:top20-overall}
\end{table}
\label{sec:wordfreq}

\subsection{Methodology}
\label{sec:wordfreq-method}

Word frequency is measured in occurrences per million tokens
(freq/M) against the raw AttaCut token total of each domain
(Table~\ref{tab:corpus-overview}).  The released lexicon uses the same
denominator, so any freq/M in Table~\ref{tab:top20-overall} can be
recomputed from the lexicon's counts.

To identify the distinctive vocabulary of each domain we use the odds
ratio, the effect-size keyness statistic 
\citep{pojanapunya-watson-todd-2018-llor}.  Log-likelihood ranks words that are frequent in the target domain even when
they are frequent everywhere; the odds ratio ranks words whose frequency
differs most between the target and the reference.  We want the second, so we
take the odds ratio and report its natural logarithm for symmetry around zero.

For each target domain $T$ and a reference corpus $R$ formed by the union
of the other three domains, the log odds ratio of word $w$ is
\begin{equation}
  \log\mathrm{OR}(w) \;=\; \ln\!\left(\frac{a \cdot d}{b \cdot c}\right),
  \label{eq:log-or}
\end{equation}
where, writing $N_T$ and $N_R$ for the total token counts of $T$ and
$R$,
\begin{align*}
  a &= \text{count of } w \text{ in } T, \\
  b &= \text{count of } w \text{ in } R, \\
  c &= N_T - a \quad (\text{all other tokens in } T), \\
  d &= N_R - b \quad (\text{all other tokens in } R).
\end{align*}  
\begin{table*}[t]
\centering
\footnotesize
\setlength{\tabcolsep}{4pt}
\begin{tabular}{rlrlrlrlr}
\toprule
& \multicolumn{2}{c}{\textbf{News}} & \multicolumn{2}{c}{\textbf{Wikipedia}} & \multicolumn{2}{c}{\textbf{Spoken}} & \multicolumn{2}{c}{\textbf{Social Media}} \\
\cmidrule(lr){2-3}\cmidrule(lr){4-5}\cmidrule(lr){6-7}\cmidrule(lr){8-9}
\textbf{Rank} & \textbf{Word} & $\log\mathrm{OR}$ & \textbf{Word} & $\log\mathrm{OR}$ & \textbf{Word} & $\log\mathrm{OR}$ & \textbf{Word} & $\log\mathrm{OR}$ \\
\midrule
1 & {\thaifont มาตรา} & 2.32 & {\thaifont แชมป์} & 2.42 & {\thaifont เนี่ย} & 4.07 & {\thaifont หุ้น} & 3.21 \\
2 & {\thaifont เชื้อ} & 2.30 & {\thaifont อังกฤษ} & 2.35 & {\thaifont ฮะ} & 3.28 & {\thaifont ธนาคาร} & 2.41 \\
3 & {\thaifont ตำรวจ} & 2.29 & {\thaifont อัลบั้ม} & 2.32 & {\thaifont แต่ว่า} & 3.25 & {\thaifont หนี้} & 2.24 \\
4 & {\thaifont คดี} & 2.16 & {\thaifont ธันวาคม} & 2.29 & {\thaifont ไอ้} & 3.11 & {\thaifont ค่ะ} & 2.19 \\
5 & {\thaifont เจ้าหน้าที่} & 2.07 & {\thaifont แข่งขัน} & 2.29 & {\thaifont ฉะนั้น} & 2.74 & {\thaifont บัตร} & 2.17 \\
6 & {\thaifont ตรวจ} & 2.06 & {\thaifont ทรง} & 2.25 & {\thaifont อ่ะ} & 2.72 & {\thaifont กู้} & 2.10 \\
7 & {\thaifont ตรวจสอบ} & 2.03 & {\thaifont เมษายน} & 2.20 & {\thaifont งี้} & 2.24 & {\thaifont ข้อความ} & 2.02 \\
8 & {\thaifont สถานการณ์} & 2.03 & {\thaifont ฟุตบอล} & 2.19 & {\thaifont เดี๋ยว} & 2.05 & {\thaifont ลอง} & 1.99 \\
9 & {\thaifont บาดเจ็บ} & 2.03 & {\thaifont พฤษภาคม} & 2.17 & {\thaifont โอเค} & 2.04 & {\thaifont จ่าย} & 1.95 \\
10 & {\thaifont ประชาชน} & 1.96 & {\thaifont กีฬา} & 2.07 & {\thaifont มัน} & 2.00 & {\thaifont ซื้อ} & 1.85 \\
11 & {\thaifont เสียหาย} & 1.94 & {\thaifont องค์} & 2.05 & {\thaifont มั้ย} & 1.94 & {\thaifont ทุน} & 1.82 \\
12 & {\thaifont มาตรการ} & 1.94 & {\thaifont ภาษา} & 1.90 & {\thaifont อาจารย์} & 1.89 & {\thaifont คง} & 1.78 \\
13 & {\thaifont นายก} & 1.93 & {\thaifont มีนาคม} & 1.89 & {\thaifont น่ะ} & 1.86 & {\thaifont แน่} & 1.78 \\
14 & {\thaifont บัญชา} & 1.92 & {\thaifont ดนตรี} & 1.86 & {\thaifont นะ} & 1.85 & {\thaifont หรอก} & 1.77 \\
15 & {\thaifont รายงาน} & 1.91 & {\thaifont ทัพ} & 1.83 & {\thaifont นึง} & 1.81 & {\thaifont เค้า} & 1.76 \\
16 & {\thaifont ระบุ} & 1.90 & {\thaifont ประวัติศาสตร์} & 1.82 & {\thaifont ใช่} & 1.68 & {\thaifont ขอบคุณ} & 1.74 \\
17 & {\thaifont เกี่ยวข้อง} & 1.90 & {\thaifont รางวัล} & 1.82 & {\thaifont อัน} & 1.51 & {\thaifont ลูกค้า} & 1.74 \\
18 & {\thaifont เหตุ} & 1.89 & {\thaifont ตะวัน} & 1.80 & {\thaifont พูด} & 1.48 & {\thaifont เยอะ} & 1.73 \\
19 & {\thaifont ยืนยัน} & 1.89 & {\thaifont ชิง} & 1.80 & {\thaifont อะไร} & 1.47 & {\thaifont สิ} & 1.71 \\
20 & {\thaifont แถลง} & 1.87 & {\thaifont ปรากฏ} & 1.78 & {\thaifont ไง} & 1.46 & {\thaifont เงิน} & 1.65 \\
\bottomrule
\end{tabular}
\caption{Top 20 words of each domain by log odds ratio against the union of the other three domains, among words occurring at least 10{,}000 times in both the target and reference and in more than ten documents of the target domain.}
\label{tab:logor-top20}
\end{table*}

\begin{figure*}[!t]
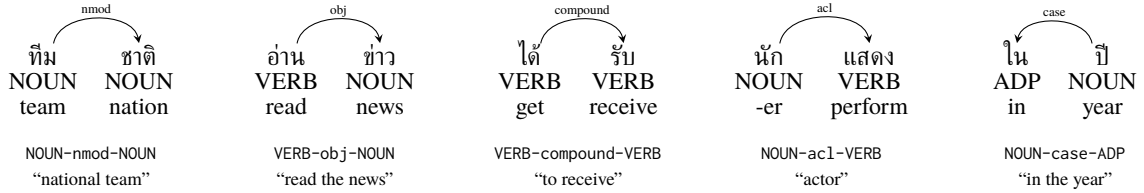

\centering
\small
\newcommand{\edgeex}[2]{\begin{minipage}[t]{0.195\textwidth}\centering #1\\[3pt]{\scriptsize #2}\end{minipage}}
\edgeex{%
\begin{dependency}[theme=simple, edge unit distance=1.5ex]
\begin{deptext}[column sep=0.22cm]
{\thaifont ทีม} \& {\thaifont ชาติ} \\
NOUN \& NOUN \\
team \& nation \\
\end{deptext}
\depedge{1}{2}{nmod}
\end{dependency}}{\texttt{NOUN-nmod-NOUN}\\ ``national team''}\hfill
\edgeex{%
\begin{dependency}[theme=simple, edge unit distance=1.5ex]
\begin{deptext}[column sep=0.22cm]
{\thaifont อ่าน} \& {\thaifont ข่าว} \\
VERB \& NOUN \\
read \& news \\
\end{deptext}
\depedge{1}{2}{obj}
\end{dependency}}{\texttt{VERB-obj-NOUN}\\ ``read the news''}\hfill
\edgeex{%
\begin{dependency}[theme=simple, edge unit distance=1.5ex]
\begin{deptext}[column sep=0.22cm]
{\thaifont ได้} \& {\thaifont รับ} \\
VERB \& VERB \\
get \& receive \\
\end{deptext}
\depedge{1}{2}{compound}
\end{dependency}}{\texttt{VERB-compound-VERB}\\ ``to receive''}\hfill
\edgeex{%
\begin{dependency}[theme=simple, edge unit distance=1.5ex]
\begin{deptext}[column sep=0.22cm]
{\thaifont นัก} \& {\thaifont แสดง} \\
NOUN \& VERB \\
-er \& perform \\
\end{deptext}
\depedge{1}{2}{acl}
\end{dependency}}{\texttt{NOUN-acl-VERB}\\ ``actor''}\hfill
\edgeex{%
\begin{dependency}[theme=simple, edge unit distance=1.5ex]
\begin{deptext}[column sep=0.22cm]
{\thaifont ใน} \& {\thaifont ปี} \\
ADP \& NOUN \\
in \& year \\
\end{deptext}
\depedge{2}{1}{case}
\end{dependency}}{\texttt{NOUN-case-ADP}\\ ``in the year''}
\caption{The five most frequent dependency-edge patterns, each shown with a
real example from the corpus.  The edge points from head to dependent and is
labeled with the Universal Dependencies relation; the pattern is written
\texttt{head-relation-dependent}.  Sources, left to right:
\texttt{news\_507:506}, \texttt{news\_5052:302}, \texttt{news\_5028:1},
\texttt{news\_5052:289}, \texttt{news\_5034:94}.}
\label{fig:edge-examples}
\end{figure*}

\input{figures/fig_bump_analysis1}
\begin{table}[!t]
\centering
\footnotesize
\setlength{\tabcolsep}{4pt}
\begin{tabular}{rlrr}
\toprule
\textbf{Rank} & \textbf{Pattern} & \textbf{Count} & \textbf{Freq/M} \\
\midrule
1 & \texttt{NOUN-nmod-NOUN} & 14{,}271{,}548 & 71{,}416.1 \\
2 & \texttt{VERB-obj-NOUN} & 13{,}929{,}524 & 69{,}704.6 \\
3 & \texttt{VERB-compound-VERB} & 9{,}735{,}805 & 48{,}718.9 \\
4 & \texttt{NOUN-acl-VERB} & 9{,}083{,}706 & 45{,}455.7 \\
5 & \texttt{NOUN-case-ADP} & 8{,}823{,}314 & 44{,}152.7 \\
6 & \texttt{VERB-advmod-ADV} & 7{,}575{,}704 & 37{,}909.5 \\
7 & \texttt{VERB-obl-NOUN} & 7{,}274{,}127 & 36{,}400.4 \\
8 & \texttt{VERB-aux-AUX} & 5{,}363{,}344 & 26{,}838.7 \\
9 & \texttt{VERB-advcl-VERB} & 5{,}023{,}108 & 25{,}136.1 \\
10 & \texttt{VERB-mark-SCONJ} & 4{,}782{,}223 & 23{,}930.7 \\
11 & \texttt{VERB-nsubj-NOUN} & 4{,}532{,}171 & 22{,}679.4 \\
12 & \texttt{VERB-conj-VERB} & 3{,}777{,}009 & 18{,}900.5 \\
13 & \texttt{NOUN-nummod-NUM} & 3{,}761{,}839 & 18{,}824.6 \\
14 & \texttt{VERB-nsubj-PRON} & 3{,}698{,}419 & 18{,}507.2 \\
15 & \texttt{NOUN-nmod-PROPN} & 3{,}548{,}850 & 17{,}758.8 \\
16 & \texttt{VERB-ccomp-VERB} & 3{,}283{,}089 & 16{,}428.9 \\
17 & \texttt{NOUN-amod-ADJ} & 2{,}860{,}823 & 14{,}315.8 \\
18 & \texttt{NOUN-compound-NOUN} & 2{,}813{,}391 & 14{,}078.5 \\
19 & \texttt{VERB-cc-CCONJ} & 2{,}499{,}434 & 12{,}507.4 \\
20 & \texttt{NOUN-compound-VERB} & 2{,}382{,}240 & 11{,}920.9 \\
\bottomrule
\end{tabular}
\caption{Top 20 dependency-edge patterns across the corpus by overall frequency per million.  Patterns are written \texttt{head-relation-dependent}.}
\label{tab:analysis1-overall}
\end{table}

\begin{table*}[!t]
\centering
\scriptsize
\setlength{\tabcolsep}{3pt}
\begin{tabular}{rlrlrlrlr}
\toprule
& \multicolumn{2}{c}{\textbf{News}} & \multicolumn{2}{c}{\textbf{Wikipedia}} & \multicolumn{2}{c}{\textbf{Spoken}} & \multicolumn{2}{c}{\textbf{Social Media}} \\
\cmidrule(lr){2-3}\cmidrule(lr){4-5}\cmidrule(lr){6-7}\cmidrule(lr){8-9}
\textbf{Rank} & \textbf{Pattern} & $\log\mathrm{OR}$ & \textbf{Pattern} & $\log\mathrm{OR}$ & \textbf{Pattern} & $\log\mathrm{OR}$ & \textbf{Pattern} & $\log\mathrm{OR}$ \\
\midrule
1 & \texttt{NOUN-appos-NUM} & 2.19 & \texttt{PROPN-appos-PROPN} & 2.93 & \texttt{PART-det-PART} & 2.80 & \texttt{ADV-punct-PUNCT} & 1.92 \\
2 & \texttt{NUM-flat-PROPN} & 1.12 & \texttt{NOUN-appos-PROPN} & 1.79 & \texttt{PART-amod-PART} & 2.77 & \texttt{VERB-obl-PUNCT} & 1.91 \\
3 & \texttt{VERB-compound-CCONJ} & 0.91 & \texttt{NOUN-nsubj-PROPN} & 1.65 & \texttt{PART-compound-PART} & 2.47 & \texttt{ADV-fixed-PART} & 1.73 \\
4 & \texttt{VERB-cc-SCONJ} & 0.83 & \texttt{PROPN-conj-NOUN} & 1.65 & \texttt{CCONJ-fixed-SCONJ} & 2.23 & \texttt{VERB-discourse-PART} & 1.72 \\
5 & \texttt{VERB-csubj-VERB} & 0.75 & \texttt{PROPN-conj-PROPN} & 1.63 & \texttt{PART-advmod-PART} & 2.17 & \texttt{ADV-advmod-PART} & 1.69 \\
6 & \texttt{NUM-advmod-ADV} & 0.69 & \texttt{PROPN-punct-PUNCT} & 1.58 & \texttt{NOUN-det-PART} & 2.16 & \texttt{VERB-dep-PART} & 1.63 \\
7 & \texttt{CCONJ-fixed-CCONJ} & 0.67 & \texttt{NOUN-conj-PROPN} & 1.56 & \texttt{VERB-advmod-DET} & 2.10 & \texttt{VERB-nsubj-PUNCT} & 1.63 \\
8 & \texttt{VERB-ccomp-SCONJ} & 0.61 & \texttt{PROPN-cop-AUX} & 1.51 & \texttt{PART-advmod-ADV} & 2.03 & \texttt{VERB-fixed-PART} & 1.58 \\
9 & \texttt{NOUN-advmod-CCONJ} & 0.59 & \texttt{NOUN-appos-NOUN} & 1.49 & \texttt{VERB-cc-PART} & 2.02 & \texttt{VERB-advmod-PART} & 1.52 \\
10 & \texttt{SCONJ-ccomp-VERB} & 0.58 & \texttt{PROPN-list-PROPN} & 1.48 & \texttt{VERB-compound-PART} & 2.01 & \texttt{NOUN-clf-NUM} & 1.47 \\
11 & \texttt{ADV-mark-PART} & 0.57 & \texttt{PROPN-nmod-PROPN} & 1.44 & \texttt{PART-fixed-PART} & 2.01 & \texttt{VERB-obj-PUNCT} & 1.47 \\
12 & \texttt{NUM-nummod-NUM} & 0.53 & \texttt{PROPN-flat-PROPN} & 1.42 & \texttt{ADV-nsubj-PRON} & 1.94 & \texttt{ADJ-punct-PUNCT} & 1.38 \\
13 & \texttt{SCONJ-flat-SCONJ} & 0.53 & \texttt{PROPN-mark-SCONJ} & 1.42 & \texttt{NOUN-det-PRON} & 1.81 & \texttt{ADV-obj-PRON} & 1.38 \\
14 & \texttt{NOUN-acl-VERB} & 0.52 & \texttt{PROPN-flat-PUNCT} & 1.37 & \texttt{VERB-aux-CCONJ} & 1.77 & \texttt{AUX-advmod-PART} & 1.33 \\
15 & \texttt{ADP-fixed-ADP} & 0.52 & \texttt{PROPN-nmod-NOUN} & 1.35 & \texttt{AUX-fixed-AUX} & 1.76 & \texttt{ADV-compound-PART} & 1.31 \\
16 & \texttt{VERB-obl-CCONJ} & 0.49 & \texttt{ADP-compound-ADP} & 1.32 & \texttt{VERB-expl-PRON} & 1.71 & \texttt{AUX-punct-PUNCT} & 1.30 \\
17 & \texttt{VERB-clf-NOUN} & 0.48 & \texttt{NOUN-flat-ADP} & 1.29 & \texttt{VERB-obj-PART} & 1.70 & \texttt{SYM-nummod-NUM} & 1.27 \\
18 & \texttt{NOUN-compound-VERB} & 0.47 & \texttt{PROPN-nmod-PUNCT} & 1.28 & \texttt{NOUN-advmod-PART} & 1.70 & \texttt{VERB-punct-PUNCT} & 1.26 \\
19 & \texttt{ADJ-obj-NOUN} & 0.47 & \texttt{PART-nmod-NOUN} & 1.27 & \texttt{PART-compound-ADV} & 1.64 & \texttt{ADV-fixed-PRON} & 1.26 \\
20 & \texttt{SCONJ-obj-NOUN} & 0.45 & \texttt{NOUN-appos-PUNCT} & 1.24 & \texttt{PART-fixed-ADV} & 1.63 & \texttt{NOUN-advmod-PART} & 1.19 \\
\bottomrule
\end{tabular}
\caption{Top 20 dependency-edge patterns of each domain by log odds ratio against the union of the other three domains. Patterns are written \texttt{head-relation-dependent} (POS of head, dependency relation, POS of dependent); only patterns with at least 10{,}000 edges in both the target and reference domains are considered.}
\label{tab:logor-analysis1}
\end{table*}

Words with $\log\mathrm{OR} > 0$ are
concentrated in the target relative to the reference; words with
$\log\mathrm{OR} < 0$ are under-represented in the target.  Because
the formula contains ``everything else'' terms on both sides ($c$ and
$d$), corpus-size differences cancel and no further normalization is
required.

Two cut-offs constrain which words are scored.  We keep only words with at
least 10{,}000 occurrences in both the target and the reference: the odds
ratio inflates for rare words, and without a floor they would dominate
the top of the ranking.  Each word must also appear in more than ten
documents of the target domain, so that no keyword comes from
a single prolific source.  \citet{pojanapunya-watson-todd-2018-llor} recommend a minimum-frequency
threshold and offer a minimum number of texts as an alternative; we apply
both.  We additionally normalize whitespace within word forms and exclude
single-character tokens, tokens containing ASCII punctuation (abbreviations
such as {\thaifont พ.ศ.}\ and {\thaifont พล.อ.}\ are not treated as lexical
words here), URL fragments, tokens consisting entirely of non-Thai,
non-Latin script, and pure-Latin tokens of fewer than three characters.  We
report the top 20 words per domain, ranked by $\log\mathrm{OR}$
descending.

These exclusions leave 354{,}430 scored word forms over
247{,}324{,}621 tokens (news 39{,}068{,}831; Wikipedia 66{,}405{,}646;
spoken 25{,}912{,}233; social media 115{,}937{,}911).  These are the totals
$N_T$ and $N_R$ of Eq.~\ref{eq:log-or}.  The keyness rankings and
cross-domain rank comparisons use these filtered forms.  The released
451{,}252-form lexicon uses a less restrictive filter: the repetition mark
{\thaifont ๆ}, for example, is among its twenty most frequent forms
(Table~\ref{tab:top20-overall}) but is excluded from keyness scoring.

\subsection{Results and Discussion}
\label{sec:wordfreq-results}

The corpus-wide word distribution follows the familiar Zipfian pattern: a
small number of forms account for a large share of all tokens.  Accordingly,
the top 20 words are predominantly function words, as is typical of frequency
analysis in a large corpus (Table~\ref{tab:top20-overall}).  Five
high-frequency verbal forms also have function-like uses: {\thaifont ได้}
can mark ability, {\thaifont ไป} and {\thaifont มา} can express deictic or
directional meanings, {\thaifont เป็น} is a copula, and {\thaifont ให้} can
function like a preposition.  The formally nominal forms {\thaifont คน} and
{\thaifont ของ} are likewise highly productive elements that combine with a
wide range of verbal material.  Thus, the upper end of the frequency list is
dominated not simply by lexical categories, but by forms with broad
grammatical and combinatory roles.
The same high-frequency forms recur in all four domains, but their rankings differ just slightly (Figure~\ref{fig:bump-top30}).

The keyness results make the domain-specific vocabulary explicit
(Table~\ref{tab:logor-top20}).  News is characterized by law, government,
and incident-reporting vocabulary, including {\thaifont ตำรวจ} ``police'',
{\thaifont คดี} ``case'', and {\thaifont มาตรา} ``legal section'', alongside
attribution verbs such as {\thaifont แถลง} ``announce'' and {\thaifont ยืนยัน}
``confirm''.  Wikipedia favors encyclopedic dates and names, with month names
and sport or entertainment terms such as {\thaifont ฟุตบอล} ``football'' and
{\thaifont อัลบั้ม} ``album''.  Spoken language is marked by conversational
particles and informal pronouns ({\thaifont เนี่ย}, {\thaifont ฮะ},
{\thaifont มัน}), while social media is associated with finance and commerce
({\thaifont หุ้น} ``stocks'', {\thaifont ธนาคาร} ``bank'', {\thaifont ซื้อ}
``buy'') as well as buyer--seller politeness ({\thaifont ค่ะ},
{\thaifont ขอบคุณ} ``thank you'').

\section{Syntactic Frequency}
\label{sec:descriptive}

\subsection{Methodology}
\label{sec:edgefreq-method}

We treat each dependency edge as an instance of a pattern: the
triplet $\langle\mathrm{head\_POS}, \mathrm{relation},
\mathrm{dep\_POS}\rangle$ of the head's part-of-speech tag, the Universal
Dependencies relation, and the dependent's part-of-speech tag. That means we do not count the subtrees. We count the edges along with the POS tags from the heads and the dependents of the edges. 

We apply the same keyness calculation method to dependency edges. The counted unit is now a syntactic pattern or a subtree instead of a word form: $a$ and $b$ are the
pattern's counts in the target domain $T$ and the reference $R$, and the totals
$E_T, E_R$ are edge counts in place of the token counts $N_T, N_R$.  As in the
lexical analysis we keep only patterns with at least 10{,}000 edges in both $T$
and $R$. Unlike word keyness calculation, we set no minimum number of documents
here.

\subsection{Results and Discussion}
\label{sec:edgefreq}

The five most frequent patterns include noun
modification, verb--object relations, and compounding
(Figure~\ref{fig:edge-examples}).  The illustrated word pairs are corpus
instances selected from the most frequent pairs for each pattern.

The 20 most frequent patterns are all well-formed and familiar Thai
constructions (Table~\ref{tab:analysis1-overall}).  This does not make every
individual automatic parse correct---the parser's attachment accuracy is below
90\%---but it provides a useful check that its most common output reflects
ordinary Thai grammar, supporting aggregate downstream analyses.  Twelve of
the patterns are headed by verbs and eight by nouns.  The leading noun pattern,
\texttt{NOUN-nmod-NOUN}, reflects the pervasive modification and compounding
of nominal expressions; the leading verbal pattern,
\texttt{VERB-obj-NOUN}, is the expected structure of a verb with a nominal
object.

Two other high-ranking patterns are particularly revealing.  The third-ranked
\texttt{VERB-compound-VERB} pattern represents serial-verb constructions,
whose 9.7M instances show that they are one of the more common and central grammtical constructions in the Thai language.  Fourth-ranked
\texttt{NOUN-acl-VERB} covers a verb modifying a noun, including relative
clauses and nominalized clauses.  Given the high corpus frequency of the
nominalizers {\thaifont การ} and {\thaifont ความ}, many instances likely
involve nominalization rather than relative clauses

The remaining frequent patterns describe similarly expected verbal and nominal
structure: adverbial modification (\texttt{VERB-advmod-ADV}), oblique nominal
dependents (\texttt{VERB-obl-NOUN}), auxiliaries
(\texttt{VERB-aux-AUX}), and subordinate clauses 
(\texttt{VERB-advcl-VERB}, \texttt{VERB-mark-SCONJ}).  Taken together, the top-20 list presents common grammatical constructions of Thai despite imperfect automatic parsing.

Looking at the top 20 syntactic patterns in each of the four domains, we see
that news and Wikipedia form one cluster, while spoken transcription and
social media form another (Figure~\ref{fig:bump-analysis1}).  News and Wikipedia
show similar rankings of dependency edges.  In spoken transcription,
\texttt{VERB-advmod-ADV}, \texttt{VERB-nsubj-PRON}, and
\texttt{VERB-aux-AUX} rank higher.  Adverbial modification covers many
syntactic phenomena.  Further inspection of the data shows that
\texttt{VERB-advmod-ADV} often involves {\thaifont ก็}, a general-purpose
discourse connective that can signal several discourse relations
\citep{prasertsom-etal-2024-thai}.  Speakers may use it more often to make
connections between sentences clear and maintain coherence despite the
natural disfluencies of speech.  Pronouns and auxiliary verbs are also more
common in spoken language than in news and Wikipedia.  This may reflect the
personal and interactive nature of speech, as well as the tendency of news
and Wikipedia to present information directly and with certainty.

\section{Conclusion}

We built ThaiTrees, a 342\,M-token corpus of Thai across four domains, to
measure how word and syntactic frequency vary between them.  By word frequency and syntactic frequency, the domains split into a formal pair (news, Wikipedia) and a conversational pair (spoken, social media).  The corpus is available in machine-readable formats, with a frequency lexicon. 

\FloatBarrier
\raggedbottom
\bibliography{custom}

\end{document}